# Synthesis and Pruning as a Dynamic Compression Strategy for Efficient Deep Neural Networks

Alastair Finlinson[1] and Sotiris Moschoyiannis[2]

Department of Computer Science,
University of Surrey, United Kingdom
[1]a.finlinson@surrey.ac.uk,
[2]s.moschoyiannis@surrey.ac.uk

**Abstract.** The brain is a highly reconfigurable machine capable of task-specific adaptations. The brain continually rewires itself for a more optimal configuration to solve problems. We propose a novel strategic synthesis algorithm for feedforward networks that draws directly from the brain's behaviours when learning. The proposed approach analyses the network and ranks weights based on their magnitude. Unlike existing approaches that advocate random selection, we select highly performing nodes as starting points for new edges and exploit the Gaussian distribution over the weights to select corresponding endpoints. The strategy aims only to produce useful connections and result in a smaller residual network structure. The approach is complemented with pruning to further the compression. We demonstrate the techniques to deep feedforward networks. The residual sub-networks that are formed from the synthesis approaches in this work form common sub-networks with similarities up to ~90%. Using pruning as a complement to the strategic synthesis approach, we observe improvements in compression.

## 1 Introduction

Modern Artificial Neural Networks are increasingly used in a wide range of domains. One emerging issue, however, is that they can often become large and take considerable time to compute results. Despite recent approaches to real-time, on-demand AI deployments [1] practical examples of deployed systems often are simply not feasible in edge and IoT use cases [2]. There are approaches to increase the compression of a deep neural network (DNN) by reducing the size of the model's parameter space and therefore, memory and storage requirements. The compression also has improvements on compute time for the device where the model is deployed.

This work explores the notion of synthesising neural networks [3]–[7]. The proposal is for a network to generate weights that are beneficial to the outcome from a very sparse network. Sparse networks are networks that have most of the connections in the network disabled for training and in deployment. Synthesis would begin with a random sub-network of a substantial architecture, that is unlikely to be optimal for the problem.

These sub-networks are viable and trainable networks, but by design are random and unlikely to be optimal in structure and weights for solving the problem.

The synthesis algorithms start generating connections, randomly or strategically, to increase the complexity of the network until the capacity threshold for generation is met. The threshold is a hyperparameter that is fixed during training.

The strategic approach to the generation of connections focuses on the high performing areas of the network (targets). These high performing connections are used to determine an origin and termination neuron for the new connections. The areas are defined as having the highest impact on the output of the network. The impact is positive when the network makes an overall improvement over a range of statistics, for example, AUC and validation accuracy [8].

Primarily, the widely used example of compression is magnitude-based pruning [9]. The compression does come at the cost of accuracy, and this is more so at higher compression ratios. The decay of accuracy is minimal until reaching a nearly completely sparse, highly compressed, network [3], [10]–[15]. This is clearly seen in [11], where compression at ~104× with a drop in accuracy of less than 4% on CIFAR-10.

The main objective is to be able to use the pruning and synthesis approaches as complementary techniques to train a model. The combination of the techniques is expected to lead to higher compression in the same models.

The key contribution of the paper is a strategic approach to reconfiguring a network, while training, so that its structure is optimal for the problem set. More specifically, we describe a synthesis approach which complements current pruning techniques and is targeted at deep neural network models. We show that our proposed approach results in a smaller residual network structure which is a more optimal configuration for the dataset, in terms of compression, while maintaining accuracy.

The remainder of this paper is structured as follows. Section 2 outlines related work. Section 3 presents the different approaches and variants we worked with. Section 4 describes the application of these approaches to a dataset and layered network. We present results in Section 5 and provide insights from these in Section 6. Section 7 concludes the paper and outlines ideas for future work.

## 2  Related Work

The works of [3]–[7] make direct use of synthesis and pruning networks to achieve compressed and accurate architectures. A direct comparison of the improvements made when using the combination of grow-prune techniques is given in [3] where the use of the techniques provide an improvement of 40% over the original accuracy and compression of eight times smaller over the prune only model of 5.7 times smaller. This is achieved using a combination of synthesis and pruning the long short-term memory cells (LSTMs) [16] during training. The synthesis and pruning processes are attributed to decreasing the inference time of the models by ~16% and error rates by 10% on average.

Xiaoliang Dai et al. [4] show that their implementation has reduced the learning costs of training the model by over 65% from scratch. They also claim that the results of the algorithm attain a higher accuracy over the baseline network. The implementation seen in [6] uses a random approach to select the next artifact to be added to the network labelled "Random Growth" and is hence closely related to our work. Shayan Hassantabar et al. also refers to a "gradient-based growth" [3] proposed initially by Dai et al. [4] and "full growth".

Neuron pruning and synthesis are techniques used in this work. The techniques for pruning and synthesis of neurons is similar to that used by Shayan Hassantabar et al. [6], Hengyuan Hu et al. [15] and Timur Ash [17], whereby, the neurons are selected at random but also by their activation values. The activation-based generation uses the fact that an active neuron is highly active; this neuron can be duplicated in the network. It is claimed that the aggressive network reduction can sustain the accuracy of the network. Our proposed approach draws from these techniques and extends them towards a strategic approach to the growth phase of the training.

In [18], a random synthesis approach on a fully connected network is proposed, where neurons are enabled until the accuracy of the network is achieved. The random generation involves a fully connected neuron being added to the network. The weights of the neuron are also fully initialised randomly. The technique described is close to the random approach to synthesis in our work.

Currently, a significant component of the compression and reduction of networks is the use of pruning. Pruning is widely used when trying to reduce complexity and maintain accuracy in the model. Pruning has been used to achieve high compression on large architectures such as with 60 times less dense on ResNet [19] and 36 times less dense MobileNet [20], while also achieving 13 times smaller with ShuffleNet [21].

Pruning is used to remove weights from a network that are contributing small value to the network. Pruning is most used to remove weights that centre around zero in their magnitude. Some of the earliest works on pruning [22], [23] show that this method of network compression can coarsely be applied and still achieve high levels of compression.

In the TensorFlow library, there are built-in pruning methods which are referred to as schedules. The available schedules are the ConstantSparsity [9] and PolynomialDecay [24]. The ConstantSparsity refers to the use of a pruning schedule that tries to prune a constant number of weights for each time it is applied to the network. This allows the user to specify a sparsity that the algorithm prunes throughout the training. The parallel drawn from TensorFlow's implementations in our work is constant sparsity. The constant sparsity model is closely replicated, in this work, to achieve similar effects as in the TensorFlow library. Constant sparsity was implemented as this is the simple case and has a lower complexity for implementation.

It transpires that the current state of the art and foundational approaches provide high levels of compression using pruning [12]–[14], [19]. There are also generative techniques [3]–[7] that show advances in the growing, referred to as synthesis in this work. The limitations in the synthesis in the above techniques show that only random techniques are directly explored. The review also shows that some explore the combination of pruning and synthesis, but this is limited. Where this work builds on compression is the use of a strategic method of synthesising networks but also then to explore the feasibility of the combination of the pruning with different approaches to synthesis. There is also a gap in the literature reviewed when looking at comparing the generated networks from different approaches. This work also touches on the network structures that form because of the different approaches. These structures may show similarity if the networks can find similar optimal solutions to the same problem and initial state.

## 3  Method
### 3.1  Approaches
There are five developed methods for compression of the models.

*Random Synthesis*
Random synthesis makes use of the initialised minimum network. From this starting point, the network is augmented with random new connections and weights until the end of the training period.

*Strategic Synthesis*
Strategic synthesis makes use of the initialised minimum network. From this starting point, the network is then augmented with new connections and weights which are selected and generated based on a ranking of current network weights. This continues until the end of the training period.

The weights are ranked according to the absolute magnitude of the tensor weights in all layers ($L$) of the network ($\gamma$). The set $\Phi$ is the top ranked magnitudes and is bound by the parameter N which is defined as the number of focal junctures.

$$X = \{\Phi :\ \text{MAX}\left(\left|\left|\gamma_L \backslash \Phi\right|\right|\right), |\Phi| < N\} \tag{1}$$

With the top N focal junctures selected, the process of generating a Gaussian distribution curve to then select the terminus of the new synthesis connection is formed.

$$f(x) = \frac{1}{\sigma\sqrt{2\pi}} * e^{\left(-\frac{(|x-\beta|-\mu)^2}{2\sigma^2}\right)} \tag{2}$$

The function uses the standard Gaussian distribution but uses the absolute distance in the potential terminus ($x$) relative the the orgin of the connection ($\beta$). The values of $\mu$ and $\sigma$ are set to 0 and 1, respectively.

*Pruning*

Pruning begins with a fully dense network with all connections and weights enabled and initialised. In this work, the pruning algorithm used is the constant pruning style. Constant pruning assumes that a fixed number of connections are removed during each step that pruning is applied.

*Random Synthesis with Pruning*

This approach begins with the minimum network. From here, the combination of random synthesis and pruning is applied on a fixed schedule.

*Strategic Synthesis with Pruning*

This approach begins with the minimum network. From here, the combination of strategic synthesis and pruning is applied on a fixed schedule.

### 3.2 Procedure

The proposed strategic approach is the first step in steering the network towards its goal of equalling or improving upon the same network. The improvement is the network's ability to solve the problem presented – measured using validation accuracy and AUC.

Validation accuracy is defined as the networks ability to correctly classify data samples as a percentage of the total number of validation samples (test samples) during training.

AUC is defined as the area in 2D space the is occupied underneath an Receiver Operating Characteristic (ROC). The ROC itself is defined as the performance of the model over varying classification thresholds.

When the combination of synthesis and pruning algorithms are applied, they are applied with the pruning first taking place then the synthesis. The ordering is used to stop newly generated connection instantly being removed from the network as there is potential for a new connection to be initialised with a low magnitude.

The steering is achieved by dynamically defining *focal junctures* (targets) in the network. These focal junctures are defined as connections in the network that are the most influential on the output (large magnitudes). The algorithm targets the highest contributing junctures, given absolute magnitudes. The aim is to reinforce the strong connections and propagate the critical features and information through the network.

The junctures have an origin (neuron) and are aware of the number of neurons in the subsequent layer. The size of the subsequent layer is used to generate a Gaussian distribution vector of the same dimensions. The vector is used to select the terminus of the new connection using the origin of the target in the subsequent layer.

The Gaussian distribution for the vector is an assumption that adjacent neurons have more relevance to one another. A neuron on the graph that exists further from the origin of the new connection is considered less likely to develop an effective pathway through the graph, and this is the initiation the vector attempts to emulate.

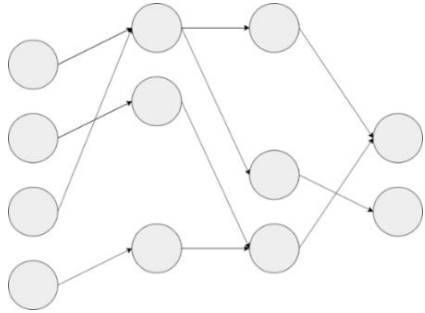

**Fig. 1.** An instance of a neural network architecture (sub-network) at a point in time during the training of the network.

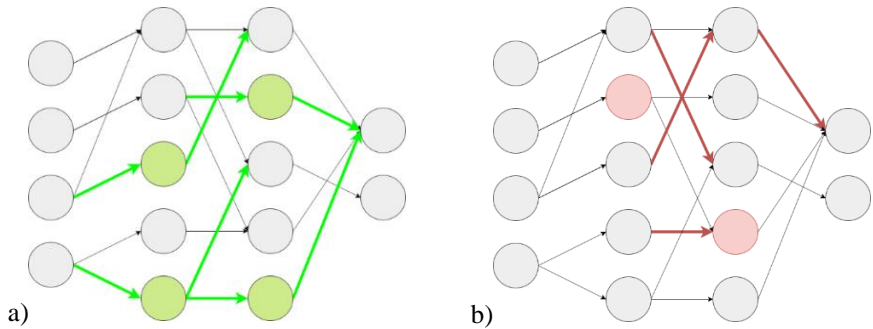

**Fig. 2. a)** Application of a synthesis algorithm; green connections and neurons are representing synthesised connections. **b)** Application of magnitude-based pruning; red connections and neurons are representing pruned connections.

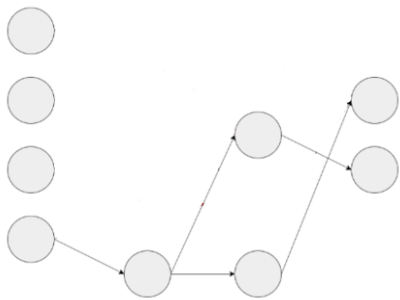

**Fig. 3.** Resulting sub-network structure after pruning and synthesis is applied

Before the network can begin training the networks require initialisation. The initialisation enables the networks from an empty architecture to be trainable, even in the infantile state. The initialisation of the network begins at each input neuron. From each neuron at most, one random path is generated through to the output. The pathways of neurons can intersect and overlap. This guarantees a connection to the output and that all inputs, at the beginning, are considered at the output.

The sub-network initialisation describes the state of a network when the synthesising processes are involved. The network is initialised to a state whereby the inputs are connected to the output through a random walk. The sub-network is initialised with this procedure such that the information about the training samples is not lost, as it could be integral to the models learning.

The connections from the initial layer to the subsequent layer are defined as having a single connection from each input neuron. From the subsequent layer, where there has been a connection made to the previous layer, a new connection can be formed. The process is a random walk from all inputs to the output along a single path. The paths can overlap or be entirely separate from the rest of the pathways. If the pathways overlap, the network only maintains a single connection at this juncture.

## 4    Problem Dataset

The dataset [25], has been selected to test the compression algorithm and is a simple binary classification problem. The data set is comprised of 303 samples, each having 13 fields, which after applying techniques such as one-hot or Boolean encoding the networks input parameters such as age, sex, chest pain type become 27 in number. The task of the neural network is to classify if the patient has heart disease or not correctly. For testing, the dataset is split into 80% for training, and the remaining 20% is used for validation. A network architecture has been selected and fixed for each of the different training methods. The architecture is shown in Figure 4 The architecture has been selected as it is a small and simple architecture, but still contains a large enough set of parameters to tune and modify. The 50 models initialised for testing were generated through TensorFlow and are the same models for all different approaches (Section 3.1).

The network used for testing is composed of an input space of 27 neurons. The subsequent layers are 16 and 8 neurons wide – Figure 4.

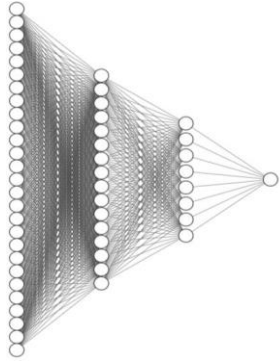

**Fig. 4.** Network Architecture Diagram for Problem Dataset.

## 5 Results

Table 1 presents the highest performing models across different pruning and synthesis thresholds. The thresholds are defined as an upper and lower limit for the algorithms to apply pruning and synthesis, respectively. The results indicate that the performance of the strategic synthesis with pruning algorithm, as averaged over the 50 test networks, has the highest performance, by ~1.5%. The AUC of the strategic synthesis with pruning varies below the dense network between 3-11%. Whereas pruning alone varies below the dense network by 10-12% - Table 3

The networks using strategic synthesis are within 0.5% of the sub-networks sparsity. This means the synthesis approach can add a minimal set of connections to the network, and this achieved an improved accuracy over the sub-networks of ~6%. With the ~3% drop in accuracy, the strategic approach alone, with the current set of parameters, is a competitive process and provides functional improvements for compression. The strategic synthesis also has the highest AUC from all results.

**Table 1.** Mean performance results for all algorithms with sparsity threshold 99%

| Model Type | Accuracy | AUC |
| --- | --- | --- |
| **Dense** | 84.83% | 78.19% |
| **Pruning** | **86.21%** | 72.52% |
| **Sub-Network** | 77.31% | 72.04% |
| **Random Synthesis** | 77.31% | 72.04% |
| **Strategic Synthesis** | 83.28% | **78.70%** |
| **Random Synthesis with Pruning** | 80.79% | 76.46% |
| **Strategic Synthesis with Pruning** | 83.62% | 75.80% |

Figure 5 shows that pruning is highly influential over the varying values for sparsity thresholding. With pruning as the benchmark, the strategic with pruning synthesis algorithm shows improvement towards the performance of the pruning as the compression ratio is increased. The strategic synthesis with pruning, as with the random synthesis and random synthesis with pruning, at lower thresholds are generating too many new connections in the network. The over synthesising of connections could be the inhibitor in early training cycles and contributing to false starts.

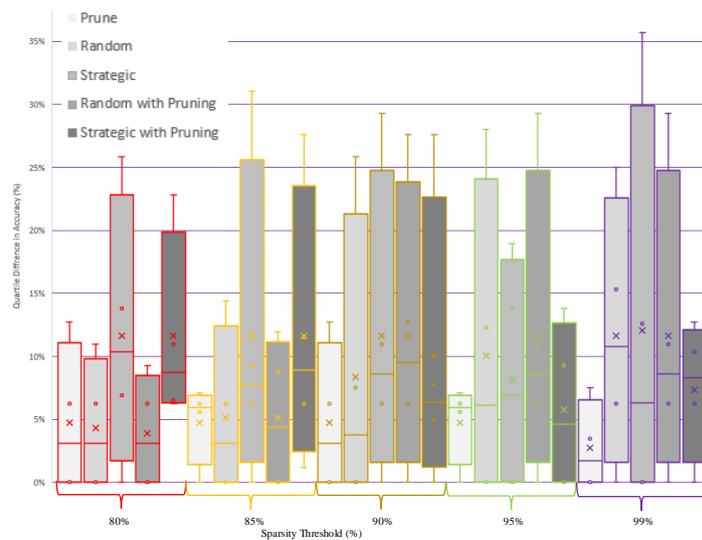

**Fig. 5.** Model validation accuracy with varying sparsity threshold. Typical values for the validation accuracy of the 50 models generated. Mean value of the samples is represented by 'x'. The 'o' represents outliers for the data. All plot plotted values are calculated using an exclusive median.

When analysing the hybrid approaches, the random synthesis approach can achieve higher compression. The higher compression, compared to the strategic synthesis, is likely to be the random weight initialisation and the instant pruning that can happen to these connections. Where the strategic is a copy of the large weight, the random can have small values approaching zero as new connection weights.

With marginally lower compression, the strategic synthesis approach can improve the performance over the random synthesis with pruning by ~3%. The performance improvement can be further enhanced, as in Table 2, where the strategic and strategic with pruning can find a solution(s) that achieves the 93.1% accuracy.

The results show that the strategic with pruning is the most capable model, achieving optimal solutions at 93.1% overall accuracy thresholds. The high performance of finding these optimal solutions shows that the problem space is better explored with this guided approach to finding a consistently optimal solution in the population of tested networks.

**Table 2.** Maximum achieved validation accuracy for each algorithm over varying sparsity thresholds from 80% to 99%

| Model Type | Sparsity Threshold | | | | |
| --- | --- | --- | --- | --- | --- |
| | 80% | 85% | 90% | 95% | 99% |
| Dense | 87.93% | 87.93% | 87.93% | 87.93% | 87.93% |
| Sub-Network | 89.66% | 89.66% | 89.66% | 89.66% | 89.66% |
| Pruning | 90.43% | 89.66% | 91.38% | 89.66% | 91.38% |
| Random Synthesis | 89.66% | 91.38% | 93.10% | 89.66% | 93.10% |
| Strategic Synthesis | 87.93% | 91.38% | **96.55%** | 91.38% | 91.38% |
| Random Synthesis with Pruning | 89.66% | **93.10%** | 93.10% | 89.66% | 91.38% |
| Strategic Synthesis with Pruning | **93.10%** | **93.10%** | 93.10% | **93.10%** | **93.10%** |

The sub-network models can perform, in the case of the maximum accuracy, better than that of the dense case. The high performance implies that one or more the randomly generated subnetworks had a structure that is more optimal than a fully dense network. The sub-network models perform within ~1% inaccuracy to the pruning networks.

With strategic synthesis with pruning performing well in searching for an optimal network, the strategic synthesis approach without pruning was able to find the most optimal sub-network for all approaches reaching an accuracy of 96.55% at a threshold of 90%.

The sub-networks are also able to achieve stability, as with the dense networks. The lower performance at a sparsity averaging 92.1% the small performance loss over the dense is marginal and when optimising for compression much more desirable.

Pruning has a lower AUC for all thresholds. Once the models set at a threshold of 85% or above, the networks exceed the performance of the pruning networks. The improvement could suggest that the other approaches better handle the changing structure of the network in training and that they are better able to persist information more effectively than pruning alone. The behaviour is particularly evident in the strategic synthesis with pruning where the algorithm is managing to outperform pruning at all thresholds.

**Table 3.** Maximum achieved validation AUC for each algorithm over varying sparsity thresholds

|  | Sparsity Threshold | | | | |
| --- | --- | --- | --- | --- | --- |
| Model Type | 80% | 85% | 90% | 95% | 99% |
| Dense | **87.04%** | **87.04%** | **87.04%** | **87.04%** | **87.04%** |
| Sub-Network | 81.56% | 81.56% | 81.56% | 81.56% | 81.56% |
| Pruning | 76.78% | 76.19% | 75.13% | 74.61% | 75.67% |
| Random Synthesis | 72.04% | 75.93% | 78.26% | 77.06% | 76.93% |
| Strategic Synthesis | 76.65% | 79.04% | 78.70% | 80.91% | 81.85% |
| Random Synthesis with Pruning | 70.65% | 76.46% | 77.19% | 76.46% | 75.20% |
| Strategic Synthesis with Pruning | 76.91% | 83.62% | 75.80% | 77.70% | 81.07% |

## 6 Further Considerations

### 6.1 Residual Networks

The networks generated using all the compression approaches form residual networks. These networks are all solving the same problem, with the same architecture and same perceived learning time, these networks converge to follow a typical core structure. The typical core structure would suggest that the optimal solution to the problem for this network can be deduced from overlaying and finding commonality in these structures.

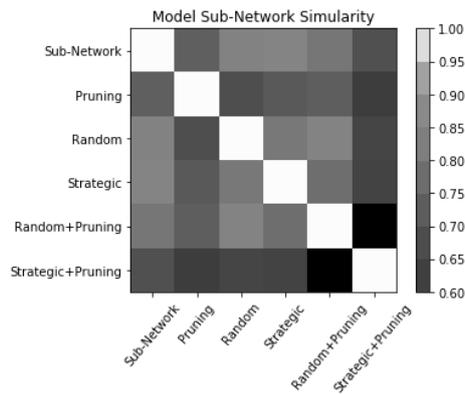

**Fig. 6.** Sub-Network Similarity Matrix

The results visualised in Fig. 6 are taken from a representative network for each compression approach; therefore, the networks selected may not fully represent and average the structure. Networks for different initialisations of the graph by TensorFlow may show that on average, the graphs share more characteristics (connections) and that a shared core network can be extracted. The extracted network would have an exceedingly high sparsity.

If a common network could be extracted, this could then be used to initialise the sub-networks. With the sub-network structure known to be performant, the network training may be reduced as well. The subnetwork could also be trained and tuned in isolation to explore its capability in the highly sparse configuration.

### 6.2 Sub-Networks

When evaluating the sub-networks used to initialise the networks for the synthesis, testing showed that there are many cases where the network can initialise and learn the problem with high accuracy. The learning of the sub-network is beneficial and has enabled the synthesis approaches to perform well.

### 6.3 False Starts

The sub-networks perform competently, on average ~77% accurate, but in the case of a false start, the accuracy can be as low as ~45%. When the algorithm does not manage to connect new useful pathways and introduces redundant connections, these networks are ineffective and useless.

The false start is propagated to the strategic synthesis as this network, in the case of the lowest-performing networks, is as low as 53.45%. This means that the strategic process was able to generate connections that would improve the network in all cases when compared to the sub-networks.

The strategic synthesis with pruning can eliminate the false starts with a minimum accuracy of 67.24%. The accuracy improvement over the sub-network at the start of training shows that the network can be improved, and that combination improves the accuracy.

### 6.4 Strategic Targets

The success rate of the networks, under synthesis, can be improved using many targets and allowing for a more significant threshold for synthesis per cycle. Increasing the number of targets does, however, have the potential to satisfy the network synthesising capacity very quickly. The earlier the network generates connections and reaches capacity, the higher the risk that the connections could stem from a connection that eventually has a weight near zero. This would make the new connections propagate low-quality information.

The increase in the number of targets should also be coupled with pruning to manage the unnecessary and rapid growth of the network and to reduce the velocity of convergence for a better-generalised model.

### 6.5 Redundancy

As the algorithms for synthesis are designed to start generating connections, randomly or strategically, there are bound to be connections generated that make no meaningful connection – suggesting that the connection either has no input or no output. With all the synthesised models, there are redundant connections. These connections, when removed post-training, would further reduce the size of the models.

### 6.6 Hybrid Scheduling

With the hybrid approaches using two sub-approaches, the scheduling and execution of these sub-processes may impact on the network. Where the approaches work synchronously, the network is managed at an approximate fixed size from the start of the processing. The network remains at the desired compression throughout training with variations in the structure. When considering an offset of different schedules for the pruning of the networks, the network can synthesise more artefacts and training them before removal. The change in the periodicity of the pruning schedule may also enable more complex pathways through the layers of the network before removing and breaking these, in-progress, pathways.

## 7 Conclusions and Future Work

Modern Artificial Neural Networks are increasingly used in a wide range of domains, from intrusion detection in cybersecurity [26], to robotics [27], image processing, or even controlling complex networks [28], [29] with applications to transport [30], business networks [31], [32] and web transactions [33], [34] among others.

In this paper we have shown that synthesising network connections has shown that it can, in some cases, perform as well as a dense or pruned network. The initialised minimum network has also shown that an optimal sub-network can be generated at random to solve this binary classification problem. The use of the sub-network generation at the start of the algorithm, before and during synthesis or pruning, provided the algorithms with a good baseline from which they successfully managed to improve upon the baseline.

The strategic synthesis with no pruning was able to improve upon the sub-network and random synthesis. The strategic algorithm in its current state is not able to improve the accuracy of the dense or pruned networks. This mismatch suggests that more parameter tuning is required. The strategic synthesis falls short of matching the dense networks by only 1.5% and the pruned networks by 3%.

When combining the strategic synthesis approach with pruning, many of the redundant connections were removed, and this resulted in improved performance, by 0.5%.

In the context of the small model used for this work, the small difference in sparsity equates to a small number of parameters. If the architecture were scaled to many thousands or millions of connections, then it would be reasonable to expect that the compression to accuracy comparison that the strategic with pruning achieves would exhibit a much more significant reduction in density.

With more rigorous tuning and improvements to the algorithm, the strategic with pruning has the potential to match the performance of the pruning process in terms of accuracy and surpass the process in terms of network sparsity (compression) and AUC.

The consistency of the strategic synthesis with pruning over varying thresholds, concerning compression and accuracy, shows that this is a stable form of generating networks. Our results show that we have produced a viable alternative and complementary technique to pruning. It manages to reproduce the same accuracy, within ~3% to 6% of the pruning approach.

There are several pathways to take this work forward. We note the following.

**Parameter Tuning**
The most interesting parameters to explore would be the initialisation strategy for artifact synthesis. The results are based on the use of the 'copy' method. This initialisation strategy has performed well and shows that the method has scope to improve. However, it would be worth exploring the 'Gaussian decay' method. This premise is because the new connections generated at the extreme of a large distribution function would be near zero and therefore, immediately pruned. This is also true for the random initialisation strategy.

**Post Training Pruning**
For this work, pruning was used as an in-process approach. However, pruning can also be used post-training to remove redundant connections. The application of this post-training pruning could further yield compression on connections that are trained out after the stop delay has occurred.

**Directed Acyclic Graph Pruning**
Directed Acyclic Graph (DAG) pruning could be devised and implemented as a post-training pruning method, that will look for pathways in the DAG that terminate before reaching the output of the network – disconnected pathways. Any path through the network that is connected from input to output is a connected pathway. The theoretical method would find all connections at the leaf of each disconnected pathway and recursively remove these connections until the pathway is no longer disconnected. It is expected that many connections are redundant, and this method of DAG pruning could remove redundant connections.

**Convolutional Neural Networks**
The future work of the strategic approaches is testing its application to the CNNs. These are classically complex and parameter dense architectures. This testing could yield results that concur with that of this work. With the much larger architectures the compression, if it follows the results of this work, could be significant. The strategic artifact selection strategies would require minimal modification to be applied to the CNN architectures.